\documentclass[10pt,twocolumn,letterpaper]{article}
\usepackage[accsupp]{axessibility}

\usepackage{authblk}
\usepackage[pagenumbers]{iccv} 

\newcommand{\mypar}[1]{\vspace{0.5mm}\noindent\textbf{#1}}

\definecolor{iccvblue}{rgb}{0.21,0.49,0.74}
\usepackage[pagebackref,breaklinks,colorlinks,allcolors=iccvblue]{hyperref}
\usepackage{comment}
\usepackage{stfloats}
\usepackage{algorithm}
\usepackage{algpseudocode}
\usepackage{float}
\usepackage[page]{appendix} 

\def\paperID{3240} 
\def\confName{ICCV}
\def\confYear{2025}

\newcommand{\OURS}{QuickSplat}
\newcommand{\SPEEDUP}{8x}

\title{\OURS: Fast 3D Surface Reconstruction via Learned Gaussian Initialization}

\author[]{Yueh-Cheng Liu}
\author[]{Lukas H{\"o}llein}
\author[]{Matthias Nie{\ss}ner}
\author[]{Angela Dai}

\affil[]{Technical University of Munich}

\begin{document}
\begin{figure}
\vspace{-4mm}
\twocolumn[{
\renewcommand\twocolumn[1][]{#1}
\maketitle

\centering
\includegraphics[width=1.02\linewidth]{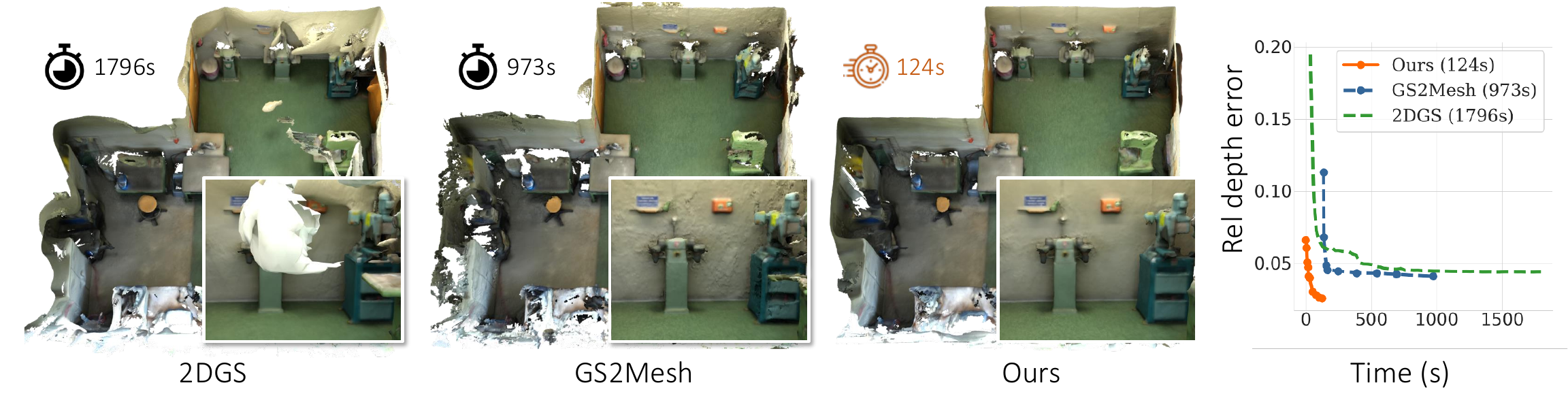}
\vspace{-6mm}

\caption{
\OURS{} performs surface reconstruction of large indoor scenes from multi-view images as input.
We learn strong priors for initialization of gaussian splatting optimization for surface reconstruction, as well as for per-iteration joint densification and gaussian updates. 
This results in high-quality mesh geometry that more accurately models flat wall structures as well as object details, while optimizing significantly faster than baselines.
}
\label{fig:teaser}
}]
\end{figure}

\begin{abstract}
Surface reconstruction is fundamental to computer vision and graphics, enabling applications in 3D modeling, mixed reality, robotics, and more. 
Existing approaches based on volumetric rendering obtain promising results, but optimize on a per-scene basis, resulting in a slow optimization that can struggle to model under-observed or textureless regions.
We introduce \OURS{}, which learns data-driven priors to generate dense initializations for 2D gaussian splatting optimization of large-scale indoor scenes.
This provides a strong starting point for the reconstruction, which accelerates the convergence of the optimization and improves the geometry of flat wall structures.
We further learn to jointly estimate the densification and update of the scene parameters during each iteration; our proposed densifier network predicts new Gaussians based on the rendering gradients of existing ones, removing the needs of heuristics for densification.
Extensive experiments on large-scale indoor scene reconstruction demonstrate the superiority of our data-driven optimization.
Concretely, we accelerate runtime by \SPEEDUP{}, while decreasing depth errors by 48\% in comparison to state of the art methods.
\end{abstract}

\section{Introduction}
\label{sec:intro}

%
%
Surface reconstruction of large, real-world scenes is a key problem in computer vision and graphics. 
Reconstructing high-quality surfaces is essential to many applications, such as creating effective virtual environments that support physical reasoning, enabling accurate simulations, and imbuing robots with crucial knowledge for navigation and interaction. 
In particular, achieving both high fidelity as well as efficient and fast reconstruction for large scenes remains a difficult problem.

%
%
Recently, 3D Gaussian Splatting (3DGS) \cite{kerbl3Dgaussians} achieves photorealistic novel-view-synthesis from multi-view images as input.
It parameterizes a scene as a set of Gaussian primitives that are initialized from SfM and subsequently optimized and densified using reconstruction losses and a differentiable volumetric renderer.
Subsequent works extend 3DGS to also obtain accurate surface reconstructions \cite{huang20242d, yu2024gaussian, guedon2024sugar,wolf2024gsmesh, chen2024pgsr}.
However, these methods typically optimize each scene separately, i.e., many iterations of gradient descent are required, which can take over 30 minutes for large-scale indoor scenes \cite{yeshwanth2023scannetpp}.
Additionally, the surface is only optimized from the observed input images, but capturing sufficiently many diverse images remains challenging for large scenes.
The resulting geometry may thus contain missing or deformed regions where there is less view coverage or texture information.


%
%
To this end, we propose a novel generalized prior for 3D surface reconstruction.
It combines high-fidelity reconstructions based on 2D Gaussian Splatting (2DGS) \cite{huang20242d} with the advantages of learning reconstruction priors from data.
Concretely, we improve the efficiency of key elements of the optimization: initialization, Gaussian updates, and densification.
This allows us to drastically accelerate the per-scene optimization time (\cref{fig:teaser}, right).
Our priors also guide the optimization towards high-quality indoor-scene geometry and thus overcome limitations stemming from insufficient observations or textureless regions (e.g., floating artifacts or non-straight wall geometry).
In comparison, our reconstructions show higher quality than state-of-the-art baselines (\cref{fig:teaser}, left).

We learn several sparse 3D CNN-based networks that jointly produce Gaussian parameters from the input posed multi-view images.
Our initializer densifies the input SfM point cloud, which enables completing large holes in unobserved or textureless regions of a scene.
We then propose a novel densification-optimization scheme, that grows new Gaussians and predicts update vectors for existing Gaussians.
Similar to gradient-descent optimizers, we iteratively improve the Gaussians by repeating this scheme multiple times.
Finally, we extract the surface from the converged 2D Gaussian primitives using TSDF fusion from rendered depth maps.
We demonstrate that our method accurately reconstructs large-scale, real-world indoor environments with arbitrary many views as input.
Extensive experiments on the ScanNet++ dataset~\cite{yeshwanth2023scannetpp} verify that \OURS{} reconstructs higher-quality geometry \SPEEDUP{} faster than baselines.


\noindent To summarize, our contributions are:
\begin{itemize}

    
    \item We propose a learned, generalized initializer network, that leverages scene priors to create effective Gaussian initializations for more efficient and accurate 3D surface reconstruction optimization, especially in under-observed or textureless areas of a scene. 
    
    \item We employ learned, generalized priors in a \textit{densification-optimization} loop, that jointly predict 2DGS updates and new densification locations.
    This heuristic-free optimization converges significantly faster and obtains more consistent geometry for large-scale scene reconstructions. 
\end{itemize}

\section{Related Work}

\subsection{Novel view synthesis}
Novel view synthesis (NVS) has received significant attention in recent years \cite{hedman2018deep, aliev2020neural, Tewari2022NeuRendSTAR, mildenhall2021nerf, kerbl3Dgaussians, barron2021mip, muller2022instant}.
NeRF~\cite{mildenhall2021nerf} renders photo-realistic images by optimizing an MLP-based scene representation with differentiable volumetric rendering.
Later, explicit or hybrid representations improved on optimization runtime  \cite{Fridovich-Keil_2022_CVPR, sun2022direct, muller2022instant, xu2022point, chen2022tensorf}.
3DGS \cite{kerbl3Dgaussians} further enables rendering at real-time rates by rasterizing explicit Gaussian primitives.
Recent methods improve the per-scene optimization spped of 3DGS by changing the underlying differentiable rasterizer, optimizer, or densification strategy \cite{durvasula2023distwar, mallick2024taming, feng2024flashgs, hollein20243dgs}.
Others learn a data-prior for sparse-view reconstruction and predict the Gaussian primitives from a feed-forward network \cite{charatan2024pixelsplat, chen2024mvsplat, xu2024depthsplat, zhang2024gs, ziwen2024long, chen2024lara}.
We similarly train a data-prior that reconstructs Gaussian primitives in a feed-forward fashion.
However, we are not limited to a sparse-view setting and focus on reconstructing better surface geometry. 

\subsection{3D reconstruction}
Reconstructing the 3D surfaces from multi-view image observations is a long-standing goal in computer vision.
Classic or neural approaches based on multi-view geometry reconstruct point clouds or mesh geometry in a multi-step pipeline based on feature matching, triangulation, and fusion \cite{schonberger2016structure, furukawa2015multi, dai2017bundlefusion, wang2024learning, bozic2021transformerfusion, zhang2022nerfusion, yao2018mvsnet, yu2020fast}.
Recently, NVS-based methods were extended to model accurate geometry and allow extracting surfaces after training through the Marching Cubes algorithm \cite{huang20242d, wang2021neus, yariv2021volume, oechsle2021unisurf, li2023neuralangelo, yu2024gaussian, guedon2024sugar}.
Our method lies in between these two directions.
We similarly train a neural network to predict surface geometry faster than optimization-based methods.
By formulating learned priors for the initialization, densification, and optimization updates of a 2DGS scene representation, we achieve improved surface reconstruction with fast runtimes.

%
%

\subsection{Meta learning}
Predicting 3D surfaces with neural networks in a feed-forward fashion typically means a single feed-forward pass produces the output.
In contrast, meta learning models the iterative optimization process with neural networks \cite{andrychowicz2016learning, wichrowska2017learned, li2017learning, finn2017model}.
Inspired by this, we frame surface reconstruction in an iterative optimization pipeline where a neural network produces the update steps.
Recent methods that model implicit functions with coordinate-based networks successfully leverage meta learning to improve the efficiency of their optimizations \cite{tancik2021learned, sitzmann2020metasdf}.
Most related to ours is G3R~\cite{chen2024g3r}, which reconstructs 3DGS primitives from multi-view RGB and dense LiDAR observations.
Their optimizer network iteratively refines the parameters of the 3D Gaussians, that are initialized from the LiDAR scan of the environment.
We leverage a similar optimizer, but additionally learn initializer and densifier networks.
This allows us to reconstruct surfaces, even for sparsely observed regions of multi-view image input. 

\section{Method}
\begin{figure*}[t]
  \centering
   \includegraphics[width=0.95\linewidth]{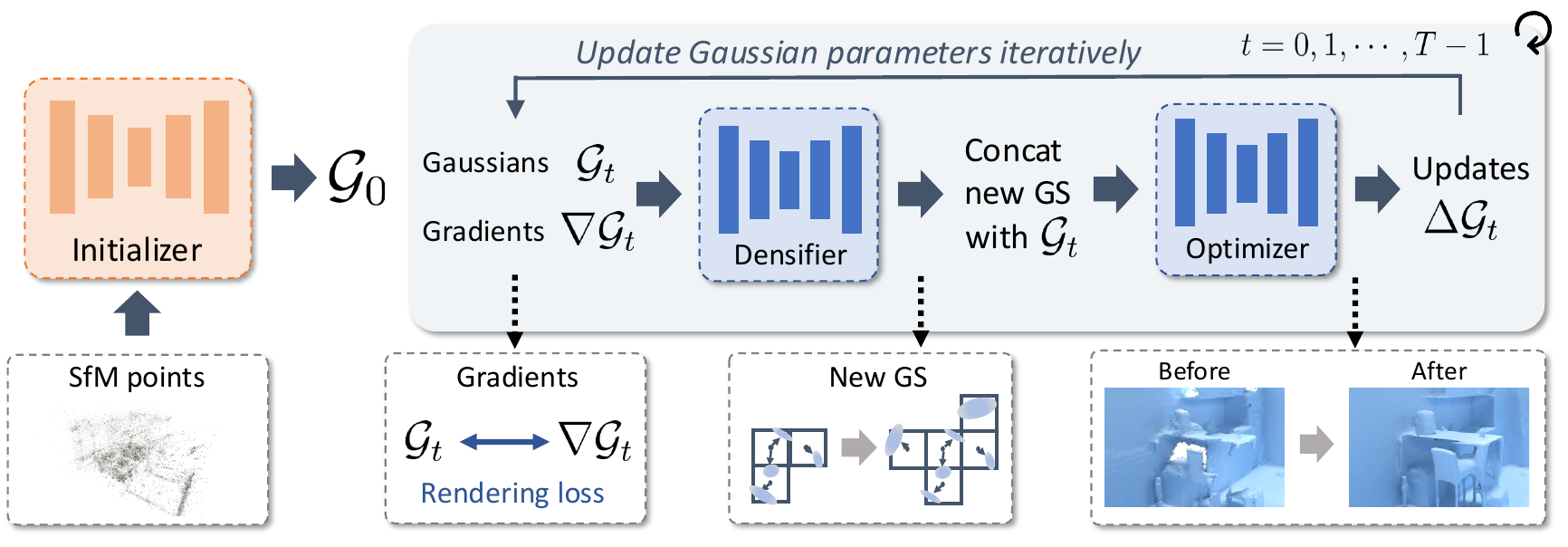}

   \caption{
   \textbf{Method overview.}
    From the SfM points of input multi-view images, our initializer network predicts an initial set of Gaussians $\mathcal{G}_0$.
    We then learn priors to improve the Gaussians during a series of optimization update steps in an iterative fashion.
    First, we calculate the current rendering gradients $\nabla\mathcal{G}_t$ using all the training images.
    The densifier network then predicts additional Gaussians  around the existing set of Gaussians $\mathcal{G}_t$.
    Finally, the optimizer network predicts an update vector $\Delta\mathcal{G}_t$, that we apply to create $\mathcal{G}_{t+1}$.
   }
   \label{fig:pipeline}
\end{figure*}

Our method reconstructs the surface of large-scale indoor scenes from posed images as input.
Specifically, we predict the attributes of 2D Gaussians \cite{huang20242d} with a novel network architecture in a feed-forward fashion (\cref{subsec:surf-rep}).
First, we initialize the scene representation from the SfM point cloud using our initializer network (\cref{subsec:init}).
Then, we incrementally improve the Gaussians in our proposed \textit{densification-optimization} loop (\cref{subsec:opt-dens}).
Concretely, we predict update vectors of scene parameters with an optimization network and grow new Gaussians with a densifier.
We summarize our method in \cref{fig:pipeline}.

\subsection{Surface Representation}
\label{subsec:surf-rep}
We adopt 2D Gaussian Splatting (2DGS) as our scene representation \cite{huang20242d}, which allows us to model high-quality surfaces.
Concretely, we use a set of Gaussian primitives $\mathcal{G} {=} \{ \mathbf{g}_i \}_{i=1}^N$ where each Gaussian $\mathbf{g}_i {\in} \mathbb{R}^{14}$ is parameterized by its 3D position, scale, and rotation, a scalar opacity and diffuse RGB color.
To render a scene from a given viewpoint, 2DGS computes the ray-splat intersection between all primitives and rays originating from the camera's image plane.
Then, a pixel color $c$ for ray $x$ is rendered using alpha-blending along the depth-sorted list of splats:
\begin{align}
\label{eq:alpha-blending}
 c(x) = \sum_{i \in \mathcal{N}} c_i \alpha_i T_i, \quad \text{with  } T_i = \prod_{j=1}^{i-1} (1 - \alpha_j),
\end{align}
where $c_i$ is the color of the $i$-th splat along the ray, $T_i$ is the accumulated transmittance, and $\alpha_i {=} o_i \mathbf{g}_i^{2D}(\mathbf{u}(x))$ the product of opacity $o_i$ and the 2D Gaussian value at the intersection point $\mathbf{u}(x)$.
The Gaussians can be optimized with rendering losses against observed image colors $C$:
\begin{align}
\label{eq:sgd-loss}
\mathcal{L}_\text{c}(x) = 0.8 || c(x) {-} C ||_1 + 0.2 (1 {-} \text{SSIM}(c(x), C)).
\end{align}
Regularizers like normal, or distortion loss \cite{barron2022mip} are also applied in addition to $\mathcal{L}_\text{c}$.
After optimization, a mesh surface is extracted by rendering depth maps from all training views and running TSDF fusion.

We propose to predict $\mathcal{G}$ with neural networks instead of optimizing the primitives directly with gradient descent.
To this end, we align the Gaussian primitives with a grid structure to facilitate predictions, similar to \cite{lu2024scaffold,roessle2024l3dg}.
Specifically, we discretize the scene into a \textit{sparse} set of 3D voxels of size $\text{v}_\text{d}$ and assign a latent feature $\mathbf{f} {\in} \mathbb{R}^{64}$ to each voxel.
Our networks predict the features of these sparse voxels.
We then decode them into $\text{v}_\text{g}$ Gaussian primitives with a small MLP.
Next, we explain the architecture of our networks in more detail.


\subsection{Initialization Prior}
\label{subsec:init}
The first step in our method is to create an initialization of all Gaussians $\mathcal{G}$.
Following 2DGS, we initialize the primitives from the SfM point cloud and voxelize them to align with our grid structure. 
This results in a sparse set of Gaussians, that does not yet represent a continuous surface, but rather contains many empty regions (e.g., SfM struggles to reconstruct points in textureless areas like walls).
We thus instead increase  surface density by predicting additional primitives around the existing ones.
Concretely, we train an initializer network $\theta_I$ that predicts additional voxel features.
Inspired by SGNN \cite{dai2020sg}, this network comprises sparse 3D convolutions in an encoder-decoder architecture.
Through a series of dense CNN blocks in the bottleneck followed by upsampling layers, the density of sparse voxels is gradually increased.
An occupancy head acts as threshold to determine if a voxel should be allocated in the next higher resolution or not.
In contrast to SGNN, which produces sparse voxel outputs, we employ a decoder MLP to interpret the densified voxel latent features as output Gaussian primitives.
This employs different activation functions for each Gaussian attribute.
The position $\mathbf{g}_c {\in} \mathbb{R}^3$ is defined relative to the voxel center $\text{v}_\text{c} {\in} \mathbb{R}^3$ as:
\begin{align}
\label{eq:pos-decode}
\mathbf{g}_c = \text{v}_\text{c} + \text{R} (2 \sigma(x) - 1)
\end{align}
where $\text{R} {=} 4 \text{v}_\text{d}$ defines the radius around the voxel in which the Gaussian primitive can live and $\sigma$ denotes the sigmoid function.
Inspired by the reparameterization trick in PixelSplat \cite{charatan2024pixelsplat}, the opacity $\mathbf{g}_o {\in} \mathbb{R}$ is the occupancy after the last upsampling layer. 
This allows rendering losses to backpropagate to the upsampling layers, controlling which points to keep in the last upsampling process.
We utilize common functions for the remaining parameters to convert the outputs into the desired value ranges.
Specifically, we utilize the softplus function for the scale parameters, sigmoid for the colors and we normalize the rotation quaternion vector.

We supervise the initializer network with multiple losses.
First, the rendering loss $\mathcal{L}_c$ (\cref{eq:sgd-loss}) provides supervision about the quality of the predicted Gaussians.
However, it does not provide signal for empty regions.
In other words, the initializer should densify the Gaussians and close any holes, but the rendering loss does not backpropagate to empty voxels.
To this end, we also supervise the occupancy of the voxel grid against ground truth geometry.
We train the network on scenes from ScanNet++ \cite{yeshwanth2023scannetpp}, which contain mesh geometries from laser scans.
This allows us to compute an occupancy loss $\mathcal{L}_\text{occ}$ before every upsampling layer of the SGNN architecture.
It comprises a binary cross-entropy loss on the occupancy of each voxel.
This provides additional signal \textit{where} new voxels should be allocated.
Furthermore, we calculate a depth loss $\mathcal{L}_{d}$, which measure the L1 distance between rendered depth and depth of the mesh for all training views.
Additionally, to make the 2D Gaussian ``disk'' more aligned with the ground-truth surfaces, we supervise the per-Gaussian normals $\mathbf{n}_\mathbf{g}$ against the mesh geometry.
The normal of a primitive is defined as the direction perpendicular to 2D disk.
We compare this against the mesh normals $\mathbf{n}_\mathbf{m}$ using cosine similarity:
\begin{align}
\label{eq:normal-loss}
\mathcal{L}_\text{n} = 1 - \mathbf{n}_\mathbf{g}^T \mathbf{n}_\mathbf{m}
\end{align}
The total loss for initializer training is then $\mathcal{L}(\theta_I) {=} \mathcal{L}_\text{c} {+} \mathcal{L}_\text{d} {+} \mathcal{L}_\text{occ} {+} 0.01\mathcal{L}_\text{n} {+} 10 \mathcal{L}_\text{dist}$ where $\mathcal{L}_\text{dist}$ denotes the distortion loss \cite{barron2022mip,huang20242d}.


\begin{figure}[t]
  \centering
  \includegraphics[width=\linewidth]{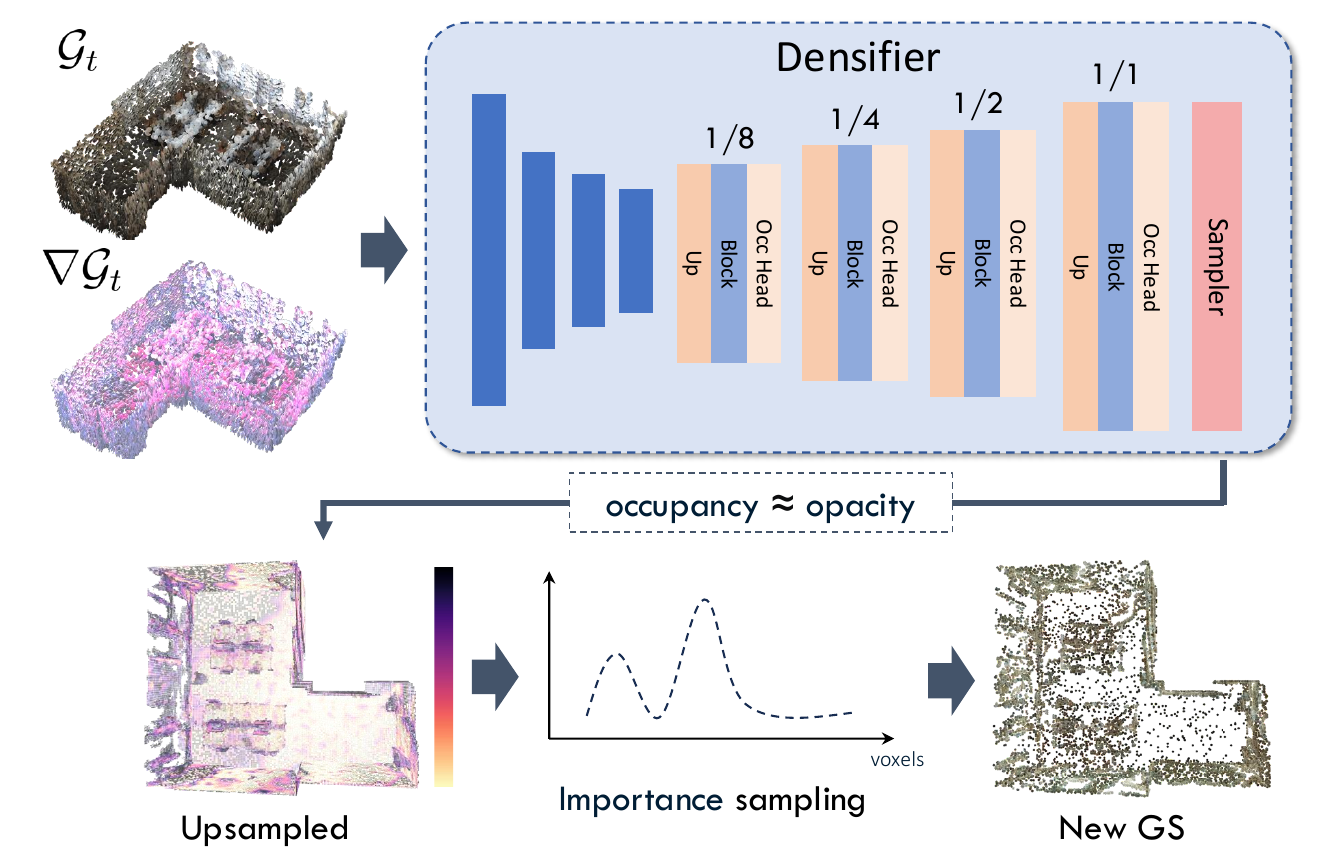}
   \caption{
    \textbf{Importance sampling of densified Gaussians.}
    Top: the densifier network predicts a pool of additional voxel features in an encoder-decoder architecture from the current Gaussians and their gradients as input.
    Bottom: to enable tractable memory during training, we select a subset of new Gaussians to densified. We apply importance sampling weighted by the occupancy predictions of the upsampled voxels. Since these occupancy values are then being used as the opacity of the densified Gaussians, this design encourages it to sample more ``solid'' Gaussians that are then passed along for the optimization updates during training. 
   }
   \label{fig:densifier}
\end{figure}

\subsection{Iterative Gaussian Optimization}
\label{subsec:opt-dens}
The initializer network predicts denser Gaussians from the SfM points as input.
While the network is supervised with a rendering loss, its input is only the geometry, i.e., the (colored) SfM points.
To further improve the surface reconstruction, we aim to also include information about the quality of the current Gaussian primitives  in the input.


Concretely, we render the training images and compute the gradients of the rendering loss~\cref{eq:sgd-loss} with respect to the latent voxel features, accumulated across multiple views. For the ease of explanation, we denote this quantity, with a slight abuse of notation, as $\nabla \mathcal{G}$. This gradient contains the signal indicating how the Gaussians should be adjusted to improve.

\mypar{Optimizer}
Inspired by G3R \cite{chen2024g3r}, we learn an optimizer network $\theta_O$, that predicts updates for all voxels from $\{\mathcal{G}, \nabla \mathcal{G}\}$ as input.
In other words, we learn the function $f_{\theta_O}(\mathcal{G}_t, \nabla \mathcal{G}_t, t) = \Delta \mathcal{G}_t$, where $\Delta \mathcal{G}_t$ denotes the predicted update for all latent voxel features.
Similar to G3R \cite{chen2024g3r}, we frame this process over multiple timesteps $t$, i.e., we iteratively calculate $\nabla \mathcal{G}_t$, predict $\Delta \mathcal{G}_t$, and update our representation as $\mathcal{G}_{t+1} {=} \mathcal{G}_t + \Delta \mathcal{G}_t$.
We utilize a sparse 3D UNet architecture \cite{choy20194d} for $\theta_O$.
We normalize its output to lie within $[-1, 1]$ to ensure that the updates do not overshoot.

\mypar{Densifier}
Our optimizer network predicts updates on a sparse voxel grid.
Although the sparse SfM point cloud gets densified by our initializer, it may still contain holes.
These holes in the surface reconstruction may never be filled by $\theta_O$, since its predicted updates can only move Gaussians around the voxel centers, but never allocate new voxels.
Empirically, we find that these holes notably reduce surface reconstruction quality (see \cref{tab:ablation}).

To this end, we introduce another learnable component, the densifier network $\theta_D$, that predicts additional voxel features in free space.
It follows the design of the initializer network $\theta_I$ with the following key differences.
First, we provide $\nabla \mathcal{G}$ and $t$ as additional inputs, i.e., we learn the function $f_{\theta_D}(\mathcal{G}_t, \nabla \mathcal{G}_t, t) {=} \mathcal{\hat{G}}_t$, where $\mathcal{\hat{G}}_t$ denotes the predicted, additional voxel features.
This enables using regions with large gradients to inform possible locations for densification.
Second, we constrain the locations of additional voxels to be neighboring to the existing ones, by using no dense blocks in the bottleneck.
We empirically find this more effective, as it better constrains where additional Gaussians should be grown.

Lastly, instead of predicting arbitrary number of new Gaussians, we perform importance sampling (see \cref{fig:densifier}) to control their number and memory usage during training. 
The key idea is to prioritize selecting new ``solid'' Gaussians to grow among other candidates generated by the decoder, since they contribute more to the surface geometry, which means selecting Gaussians with higher opacity values.
By leveraging the reparameterization trick (\ie, interpreting occupancy as opacity), we can sample additional voxels weighted by the occupancy prediction after the last upsampling layer.
We define the number of voxels that can be added in the current iteration as $n(t) {=} s/(2^t)$ with $s {=} 20\text{K}$.
In other words, we densify more for earlier timesteps and then gradually reduce the number of new voxels.
 During test time, we simply select top $n(t)$ voxels from the occupancy prediction.

The benefit of this method is that it avoids designing a densification strategy based on heuristics~\cite{kerbl3Dgaussians,mallick2024taming,rota2024revising,fang2024mini}.
By training the densifier network end-to-end with the optimizer, we instead learn to map the current state of Gaussians and their gradients into new, high-contribution Gaussians.

\mypar{Densification-Optimization Loop}
We utilize both networks, $\theta_D$ and $\theta_O$, in our proposed densification-optimization loop, that grows and improves the latent voxel features over multiple timesteps.
First, we decode the voxels into Gaussian parameters, render all training images, and calculate $\nabla \mathcal{G}_t$.
Then, the densifier predicts additional voxel positions $\mathcal{\hat{G}}_t$.
We concatenate the existing and novel voxel features as $\mathcal{\bar{G}}_t {=} \mathcal{G}_t \cup \mathcal{\hat{G}}_t$ and initialize the gradient of novel voxels with zeros to similarly obtain $\nabla \mathcal{\bar{G}}_t$.
The optimizer then predicts the update $\Delta \mathcal{\bar{G}}_t$ and we obtain $\mathcal{G}_{t+1} {=}  \mathcal{\bar{G}}_t {+} \Delta \mathcal{\bar{G}}_t$.

\mypar{End-to-End Training}
We jointly train the densifier and optimizer networks in our second training stage.
The initializer network $\theta_I$ remains frozen during this stage.
We similarly calculate $\mathcal{L}_\text{occ}(\theta_D)$ on the intermediate upsampling layers of the densifier.
Additionally, we decode the updated voxel features $\mathcal{G}_{t+1}$ into Gaussians and calculate the rendering loss $\mathcal{L}(\theta_O) {=} \mathcal{L}_\text{c} {+} \mathcal{L}_\text{d} {+} 10 \mathcal{L}_\text{dist}$.
We run the densification-optimization loop for $T {=} 5$ timesteps and calculate the losses after each timestep.
Similarly to G3R \cite{chen2024g3r}, we detach the gradient of the losses for the subsequent timesteps, i.e., we optimize them separately.

\begin{table*}[tp]
  \centering
  \begin{tabular}{@{}lccccccc@{}}
    \toprule
    Method & Abs err$\downarrow$ & Acc (2cm)$\uparrow$ & Acc (5cm)$\uparrow$ &  Acc (10cm)$\uparrow$ & Chamfer$\downarrow$ & Time$\downarrow$ \\
    \midrule
    SuGaR~\cite{guedon2024sugar}  & 0.2061    & 0.1157    & 0.2774    & 0.4794 & 0.2078 & 3130s \\
    2DGS~\cite{huang20242d} &  0.1127 & 0.4021 & 0.6027 & 0.7422 & 0.2420 & 1796s \\
    PGSR~\cite{chen2024pgsr} & 0.2325 & 0.4795 & 0.6407 & 0.7496 &  0.2228 & 2593s \\
    GS2Mesh~\cite{wolf2024gsmesh} & 0.1212 & 0.4028 &0.6039 & 0.7406 &  0.2012  & 973s \\
    MonoSDF~\cite{yu2022monosdf} & \textbf{0.0569} & \underline{0.5774} & \underline{0.8006} & \underline{0.8850} & \underline{0.1450} & $>$10h \\
    \midrule
    Ours (w/o opt) & 0.0732 & 0.5263 & 0.7674 & 0.8583 & 0.1461 & \textbf{26s} \\
    Ours (w/ opt)  &  \underline{0.0578} & \textbf{0.5783} & \textbf{0.8035} &  \textbf{0.8887} & \textbf{0.1347} & \underline{124s} \\
    \bottomrule
  \end{tabular}
  \caption{
  \textbf{Quantitative comparison against baselines.}
  We compare the quality and optimization runtime of our reconstructed surfaces against baseline methods, and show averaged results on the test scenes in ScanNet++~\cite{yeshwanth2023scannetpp}.
  Both our method without post-training (``w/o opt'') and with additional SGD iterations (``w/ opt'') obtain better geometry while achieving orders of magnitude faster runtime.
  }
  \label{tab:main}
\end{table*}

\section{Experiments}

\begin{figure*}[t]
  \centering
   \includegraphics[width=\linewidth]{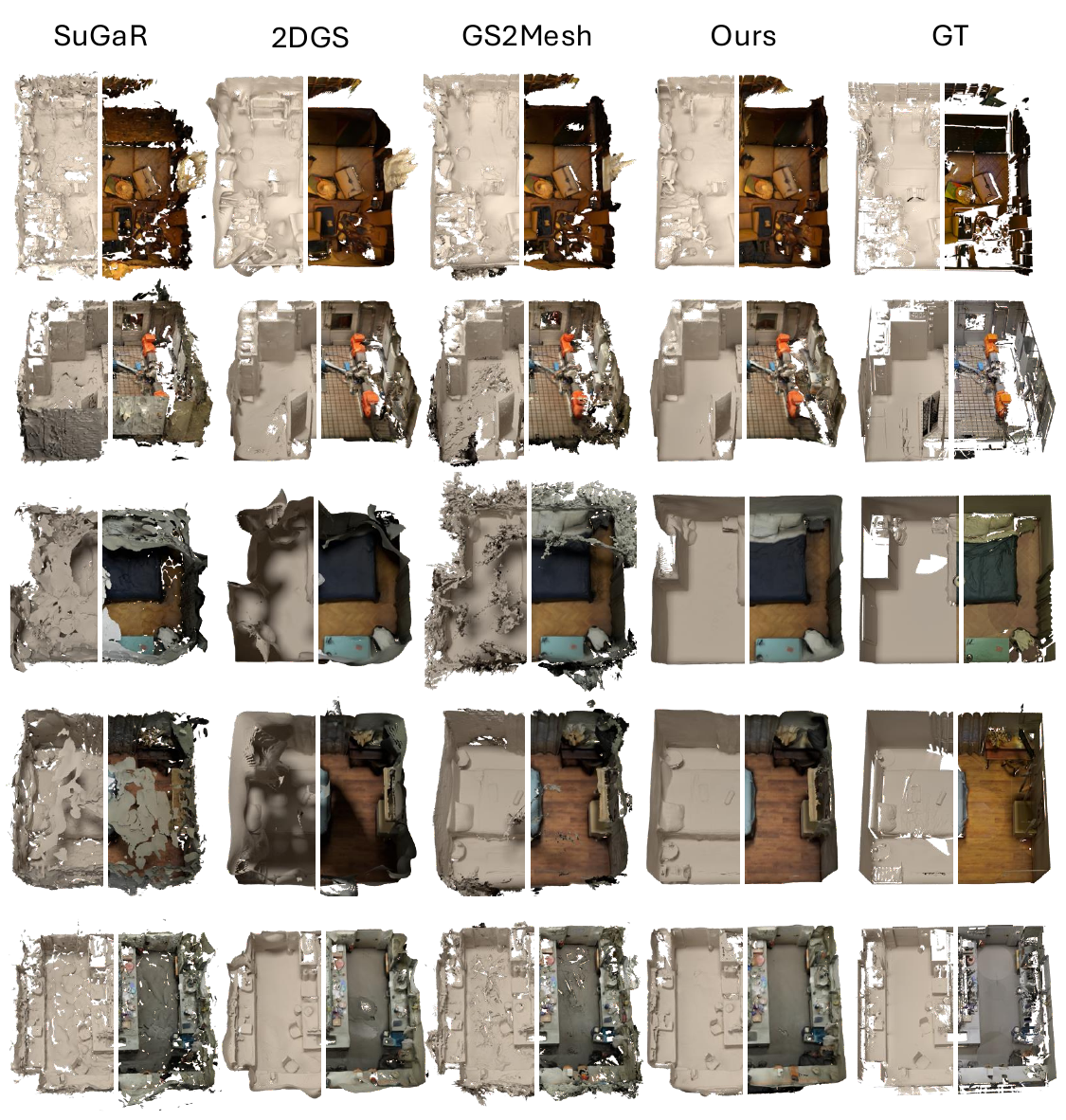}
   \caption{
   \textbf{Qualitative comparison against baselines.}
    We show top-down views of reconstructed mesh geometries (with and without vertex colors) in comparison to the ground-truth meshes of ScanNet++~\cite{yeshwanth2023scannetpp}.
    Our method more accurately models flat wall structures and objects details, while producing fewer floating geometry artifacts.
   }
   \label{fig:main-result}
\end{figure*}

\begin{figure}[t]
  \centering
  \includegraphics[width=\linewidth]{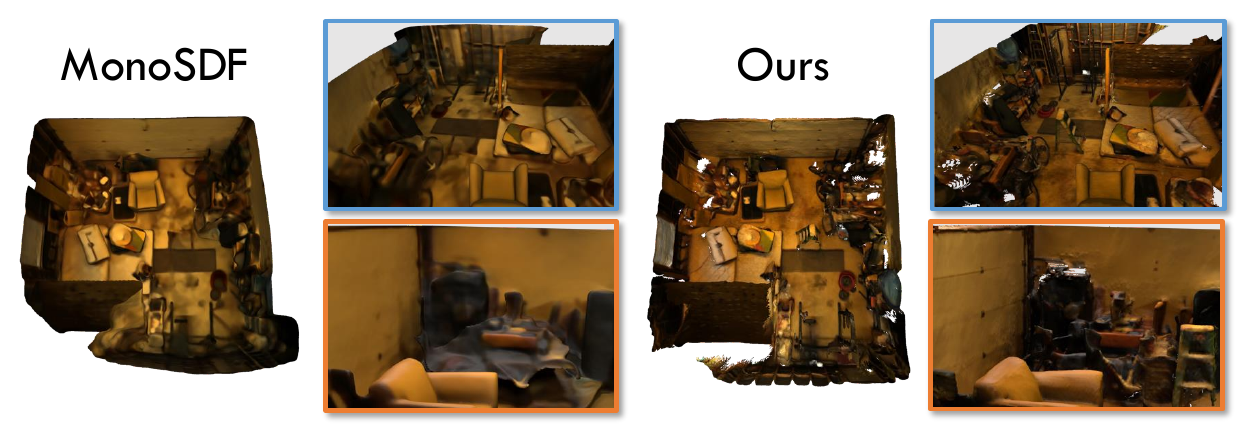}
  \vspace{-6mm}
   \caption{
    \textbf{Qualitative comparison between MonoSDF~\cite{yu2022monosdf} and ours.} Our QuickSplat achieves faster reconstruction and retains more fine details. For example, MonoSDF fails to reconstruct the ladder since the pretrained depth/normal estimator misses it.
}
   \label{fig:monosdf}
\end{figure}


\mypar{Dataset}
We train and evaluate our model on the ScanNet++ dataset \cite{yeshwanth2023scannetpp}.
We utilize 902 indoor scenes for training, after filtering out some scenes with incomplete wall structures or very large bounding extents.
We train on the undistorted DSLR images at $360 \times 540$ resolution and double the resolution during evaluation to process images at $720 \times 1080$.
We evaluate our method on 20 unseen test scenes and report averaged metrics.


\begin{table*}
  \centering
  \begin{tabular}{@{}ccccccc@{}}
    \toprule
     \# & Initializer & Densifier & Optimizer & Abs err$\downarrow$ & Chamfer$\downarrow$ & \#Gaussians \\
    \midrule
    (a) &   -        &   -     & $\surd$  & 0.1332 & 0.2881 & 47k  \\
    (b)&   occ only &   -     & $\surd$  & 0.0897 & 0.2095 & 276k \\
    (c)&   occ only & $\surd$ & $\surd$  & 0.0844 & 0.2038 & 361k \\
    (d)&  $\surd$  &     -   & $\surd$  & 0.0581 & 0.1374 & 184k \\
    (e)&  $\surd$  & $\surd$ & $\surd$  & \textbf{0.0578} & \textbf{0.1347} & 251k \\
    \bottomrule
  \end{tabular}
  \caption{
  \textbf{Ablation study.}
  We ablate the impact of our learned priors for initialization, densification, and optimization updates.
  Only using our optimizer network does not increase the number of Gaussians and thus struggles to model continuous surfaces (a).
  Densifying the SfM point cloud, instead of predicting initial Gaussian parameters results in less accurate geometry (``occ only'').
  The densifier further increases the number of Gaussians, which helps to further improve surface quality.
  }
  \label{tab:ablation}
\end{table*}

\mypar{Implementation Details}
We set $\text{v}_\text{d} {=} 4$cm and predict $\text{v}_\text{g} {=} 2$ Gaussians per voxel in all our experiments.
The initializer and densifier network use four up/downsampling layers, and the optimizer UNet architecture follows G3R \cite{chen2024g3r}.
In total the networks have around 68M parameters.
We set the learning rate to $1\mathrm{e}{-4}$ and train the networks for 3 days on a single Nvidia RTX A6000.
In each iteration, we accumulate the gradients $\nabla \mathcal{G}$ from 100 training images.
After running our iterative optimization for $t {=} 5$ timesteps, we optionally refine the Gaussians for another 2000 steps of gradient descent (without adaptive density control).
We denote results of our method as ``w/ opt'' that use this refinement and as ``w/o opt'' if they do not use it.



\mypar{Baselines}
We compare our method with several recent 3D surface reconstruction approaches: SuGaR~\cite{guedon2024sugar}, 2DGS~\cite{huang20242d}, GS2Mesh~\cite{wolf2024gsmesh}, PGSR~\cite{chen2024pgsr}, and MonoSDF~\cite{yu2022monosdf}.
SuGaR optimizes 3DGS~\cite{kerbl3Dgaussians} and regularizes the Gaussians to align with surfaces, followed by mesh extraction using Poisson surface reconstruction.
Similarly, GS2Mesh also begins by optimizing 3DGS, then renders stereo pairs and predicts depth maps using an off-the-shelf stereo depth estimator (i.e., DLNR~\cite{zhao2023high}).
2DGS optimizes and densifies flat 2D Gaussians using gradient descent in a per-scene optimization lasting 30K iterations.
PGSR renders unbiased depth maps from flattened 3D Gaussians and introduces both single-view and multi-view regularization losses to improve geometric reconstruction.
MonoSDF leverages monocular depth and normal priors from a pre-trained model~\cite{eftekhar2021omnidata} to optimize the signed distance field (SDF) using differentiable volumetric rendering.

\mypar{Metrics}
To evaluate the quality of the reconstructed geometry, we measure the error between rendered depth and the ground-truth depth maps of ScanNet++ testing frames.
We calculate the absolute error, as well as the accuracy within different thresholds (2cm, 5cm, 10cm).
We also calculate the Chamfer distance between the predicted and ground-truth mesh vertices.
We crop the predicted vertices outside of ground-truth bounding box to prevent from penalizing false negative predictions outside of windows.
Additionally, we report the optimization runtime in seconds.


\begin{figure}[tp]
  \centering
   \includegraphics[width=\linewidth]{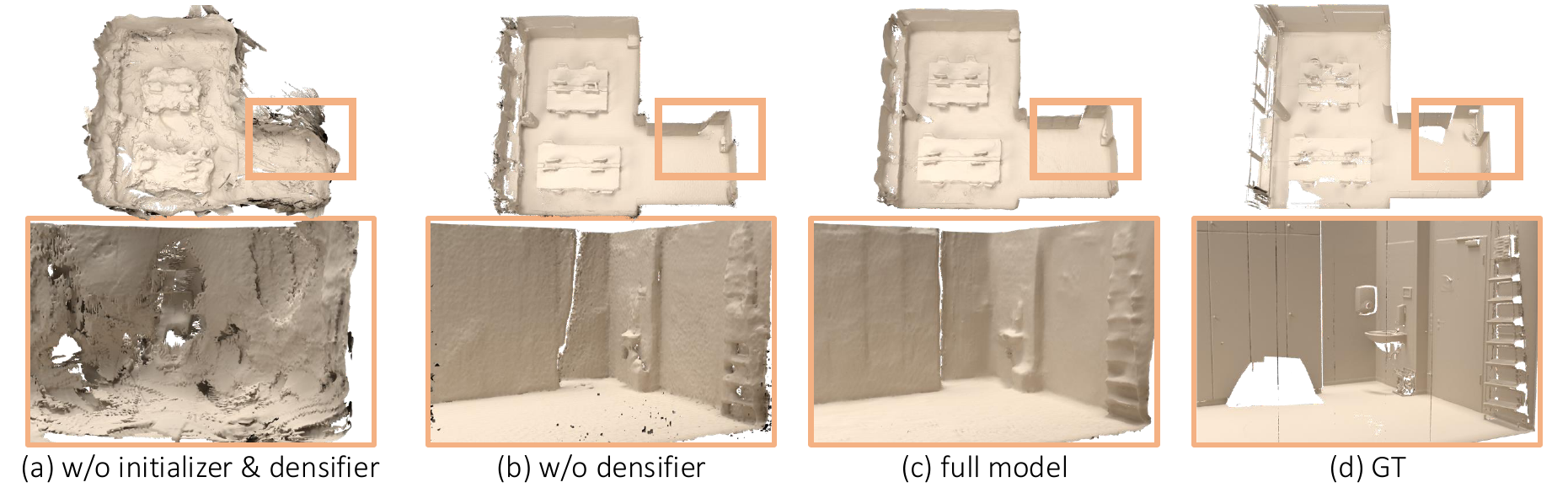}
   \caption{
   \textbf{Visualization of ablations.}
   (a) Without our initializer and densification priors during optimization, surface reconstruction of untextured regions such as walls is challenging due to the lack of SfM points.
   (b) With our initializer but without densification predictions, reconstruction improves, but maintains smaller-scale artifacts around regions with holes or excess geometry. 
   (c) Our full model produces a robust reconstruction of the scene while achieving fast optimization runtimes. 
   }
   \label{fig:ablation}
\end{figure}

\subsection{Comparison to State of the Art}
Experimental results on ScanNet++ are shown in \cref{tab:main} and \cref{fig:main-result}.
In general, our proposed \OURS{} achieves better performance: it reconstructs scenes with cleaner structures and flat surfaces that matches the ground truth compared to the baselines while maintaining similar level of details.
Most importantly, via the learned initializer and densifier-optimizer, our method converges much faster.


Compared to SuGaR~\cite{guedon2024sugar}, 2DGS~\cite{huang20242d}, and PGSR~\cite{chen2024pgsr}, which are optimized per scene using rendering losses, our method additionally leverages learned geometry priors from data.
This is particularly useful in indoor environments, which often contain large textureless regions (e.g., white walls).
Methods that rely solely on photometric errors may struggle in these areas (see \cref{fig:main-result} for examples), and therefore produce curved surfaces and floating artifacts. In contrast, our learned initializer network predicts a relatively dense Gaussian initialization, effectively addressing the issue of missing SfM points in textureless regions.

Similar to our approach, GS2Mesh~\cite{wolf2024gsmesh} also utilizes geometry priors learned from data (\ie, from a pretrained stereo estimator). However, since synthesized views from 3DGS may not be as realistic as actual images, a noticeable domain gap exists for stereo depth estimators when the renderings contain noise. This can lead to high-frequency artifacts in the reconstructed geometry (see \cref{tab:main}). In contrast, our method learns scene priors directly in the Gaussian representation space, allowing it to perform more robustly under such conditions.

Our method matches MonoSDF in accuracy while running substantially faster. Moreover, unlike MonoSDF—which sometimes produces overly smooth surfaces and misses fine structures (e.g., the ladder in \cref{fig:monosdf}) when the monocular depth or normal estimator fails to capture thin geometry—our approach better preserves fine details.

After the iterative Gaussian optimization with our learned prior networks (w/o opt), \OURS{} can further improve the geometry quality by running short iterations of post-training (w/ opt).
It is especially beneficial for improving fine details, which results in 5\% increase in Acc (2cm). 
Nevertheless, even without additional per-scene optimization, we significantly reduce the runtime while maintaining state-of-the-art performance.



\subsection{Ablations}

Our method combines three data prior networks in an iterative fashion (\cref{fig:pipeline}).
We demonstrate the importance of each individual component.

\mypar{Only optimizer network} 
Without the initializer and the densifier, the optimizer network relies solely on sparse SfM point clouds. 
Since it has no ability to generate additional Gaussians, it is hard to accurately represent the surface geometry.
As shown in \cref{tab:ablation}~(a), the number of Gaussians remains too low, making the performance significantly worse.
An example of the output can be seen in \cref{fig:ablation}~(a), where it struggles to model reasonable surfaces (e.g., the walls are bent and noisy).


\mypar{Importance of the initializer}
To evaluate the effect of the initializer, we train a network that predicts dense point clouds from SfM points and supervise it only with $\mathcal{L}_\text{occ}$ (referred to as ``occ only'' in \cref{tab:ablation}).
In contrast, our full initializer predicts dense points with 2DGS attributes directly.
Since the occupancy-only initializer has no knowledge over the Gaussian representation and their rendering quality, the performance is worse than our proposed initializer (see (b) \vs (d) and (c) \vs (e) in \cref{tab:ablation}).


\mypar{Importance of the densifier}
We compare the reconstruction performance with and without the proposed densifier.
As can be seen in \cref{fig:ablation} (b) \vs (c), even with the dense initialization, the densifier is able to fill the remaining holes and predict new Gaussians adaptively, based on the current state of the Gaussian representation and their gradients.
Therefore, it further improves the geometry details, such as the stairs at the corner.
As shown in \cref{tab:ablation} ((b) \vs (c) and (d) \vs (e)), this results in a noticeable performance improvement.

\mypar{Gaussian attributes initialization} 
We further investigate the effect of predicting only a subset of the Gaussian attributes in our initializer.
It can be seen in \cref{tab:attribute} that initializing each attribute of the Gaussian representation contributes to improvements.
Therefore, we used all four additional Gaussian attributes for \OURS{} (next to predicting the position).





\subsection{Limitations}
Our method accelerates optimization runtime by \SPEEDUP{} and obtains more accurate surface reconstructions from posed images in comparison to baselines.
However, some drawbacks remain.
First, our method struggles with mirror reflections, since the photometric loss encourages to reconstruct the reflected geometry behind the mirror, which leads to noisy artifacts.
Second, we assume static environments and therefore cannot reconstruct dynamic scenes (e.g., people walking inside of a room).
Lastly, even though we significantly reduce optimization runtime, our method does not yet reconstruct in real-time, but could be integrated with recent SLAM-based approaches \cite{matsuki2024gaussian, keetha2024splatam, yan2024gs}. 


\begin{table}[t]
  \footnotesize
  \centering
  \begin{tabular}{@{}cccccc@{}}
    \toprule
    Color & Opacity & Scales & Rotation & Abs err$\downarrow$ & Chamfer$\downarrow$ \\
    \midrule
    $\surd$ &           &          &          & 0.0627 & 0.1500     \\
    $\surd$ & $\surd$   &          &          & 0.0600 & 0.1421   \\
    $\surd$ & $\surd$   & $\surd$  &          & 0.0590 & 0.1402   \\
    $\surd$ & $\surd$   & $\surd$  & $\surd$  & \textbf{0.0578} & \textbf{0.1347}   \\
    \bottomrule
  \end{tabular}
  \caption{
  \textbf{Initializer output ablation study.}
  We evaluate the impact of predicting different Gaussian attributes from the SfM point cloud with our initializer network.
  Predicting all attributes results in the best final surface reconstruction quality.
  }
  \label{tab:attribute}
\end{table}

\section{Conclusion}
We have presented \OURS{}, which learns several data priors to perform surface reconstruction optimization of large indoor scenes from multi-view images as input.
By framing the optimization with learned prior networks for initialization, densification, and optimization updates of 2D Gaussian splats, we significantly accelerate surface reconstruction speed by \SPEEDUP{} in comparison to baselines.
Furthermore, we demonstrate that incorporating data-priors helps reduce artifacts caused by insufficiently many observations or textureless areas, that typically occur in large-scale scene reconstructions.
That is, our initializer network densifies the input SfM points by exploiting learned geometry priors (e.g., flat wall structures).
Then, our proposed \textit{densification-optimization} loop refines the Gaussian attributes through a series of predicted update steps.
Overall, we believe that the ability to utilize data-priors for fast and state-of-the-art reconstructions will open up further research avenues and make surface reconstructions more practical across a wide range of real-world applications.
\section*{Acknowledgements}
This work was supported by the ERC Starting Grant SpatialSem (101076253), the ERC Consolidator Grant Gen3D (101171131), and the German Research Foundation (DFG) Research Unit ``Learning and Simulation in Visual Computing.''

{
    \small
    \bibliographystyle{ieeenat_fullname}
    \bibliography{main}
}

\clearpage

\renewcommand{\appendixpagename}{\Large{Appendix}} 
\begin{appendices}

In the appendix we provide further details of the method and the baselines in \cref{sec:more-details}, more surface reconstruction results on ScanNet++ and other datasets like Arkitscenes in \cref{sec:more-result}, more ablations in \cref{sec:more-ablations}.

\section{More Implementation Details}\label{sec:more-details}

\paragraph{SuGaR.}
We follow the official code that optimizes vanilla 3DGS for 7,000 iterations and refine for 15,000 iterations to get the best quality mesh. Depth-normal consistency (dn\_consistency) is used as the regularization objective.

\paragraph{2DGS.} We follow the official code and optimize the scene for 30,000 iterations, using the same hyper-parameters such as the learning rates and the number of iterations for pruning and densification; we only optimize the RGB color of the Gaussians instead of the spherical harmonics.

\paragraph{GS2Mesh.} We follow the official code and optimize vanilla 3DGS for 30,000 iterations. The pretrained stereo estimation model from DLNR~\cite{zhao2023high} that is trained on Middlebury is used to extract stereo depth, with 0.1m as the stereo baseline. Since we work on scene-level datasets, the object masks are ignored.

\paragraph{MonoSDF.} We follow the official code and use MLP as the scene representation. We use 
the Omnidata~\cite{eftekhar2021omnidata} to extract the depth and normal of the training images, and both depth and normal losses are used for the optimization. The model is optimized for 1,000 epochs.

\paragraph{PGSR. (Chen et al. 2024)} We uses the official code and optimize the scenes for 30,000 iterations, with single view and multi-view regularization loss after 7,000 iterations. Exposure compensation is not used as ScanNet++ has fixed camera exposure.

\paragraph{QuickSplat.}
We provide the pseudo code of the optimization process of QuickSplat in \cref{alg:quicksplat}.

\begin{algorithm}[thb]
\caption{The optimization process of QuickSplat}\label{alg:quicksplat}
\begin{algorithmic}
\State $\mathcal{P}$: SfM points
\State $f_{I}$: initializer network
\State $f_{D}$: densifier network
\State $f_{O}$: optimizer network
\\
\State $\mathcal{G}_0 \gets f_{I}(\mathcal{P})$
\For{$t = 0$ \text{to} $T-1$}
    \State $\nabla \mathcal{G}_t \gets 0$
    \ForAll{images}
        \State $L \gets$ rendering loss of the image
        \State $\nabla \mathcal{G}_t \gets \nabla \mathcal{G}_t + \frac{\delta L}{\delta  \mathcal{G}_t}$
    \EndFor
    \State $\mathcal{\hat{G}}_t \gets f_D(\mathcal{G}_t , \nabla \mathcal{G}_t, t)$
    \State $\mathcal{\bar{G}}_t \gets \mathcal{G}_t \cup \mathcal{\hat{G}}_t$  \Comment{Concatenate the new GS}
    \\
    \State $\nabla \mathcal{\bar{G}}_t \gets  \nabla \mathcal{G}_t \cup \mathbf{0}$
    \State $\Delta \mathcal{\bar{G}}_t \gets f_O ( \mathcal{\bar{G}}_t , \nabla \mathcal{\bar{G}}_t , t)$
    \State $\mathcal{G}_{t+1} \gets \mathcal{\bar{G}}_t + \Delta \mathcal{\bar{G}}_t$ \Comment{Update the parameters}
\EndFor

\end{algorithmic}
\end{algorithm}

\section{Additional Results}\label{sec:more-result}

\paragraph{Generalization.} 
To demonstrate the generalization ability of our method, we run QuickSplat trained on ScanNet++ directly on other indoor datasets, such as ARKitScenes~\cite{dehghan2021arkitscenes} and Mip-NeRF 360~\cite{barron2022mip}, without any additional fine-tuning.



We process the ARKitScenes dataset following the same procedure as ScanNet++, obtaining the SfM point clouds and the alignment between camera poses and the ground-truth mesh. For Mip-NeRF 360 (Room), we restore the absolute scale of the official COLMAP point cloud and poses using a monocular metric depth estimator~\cite{depth_anything_v2}.

\begin{figure}[h]
  \centering
  \includegraphics[width=0.9\linewidth]{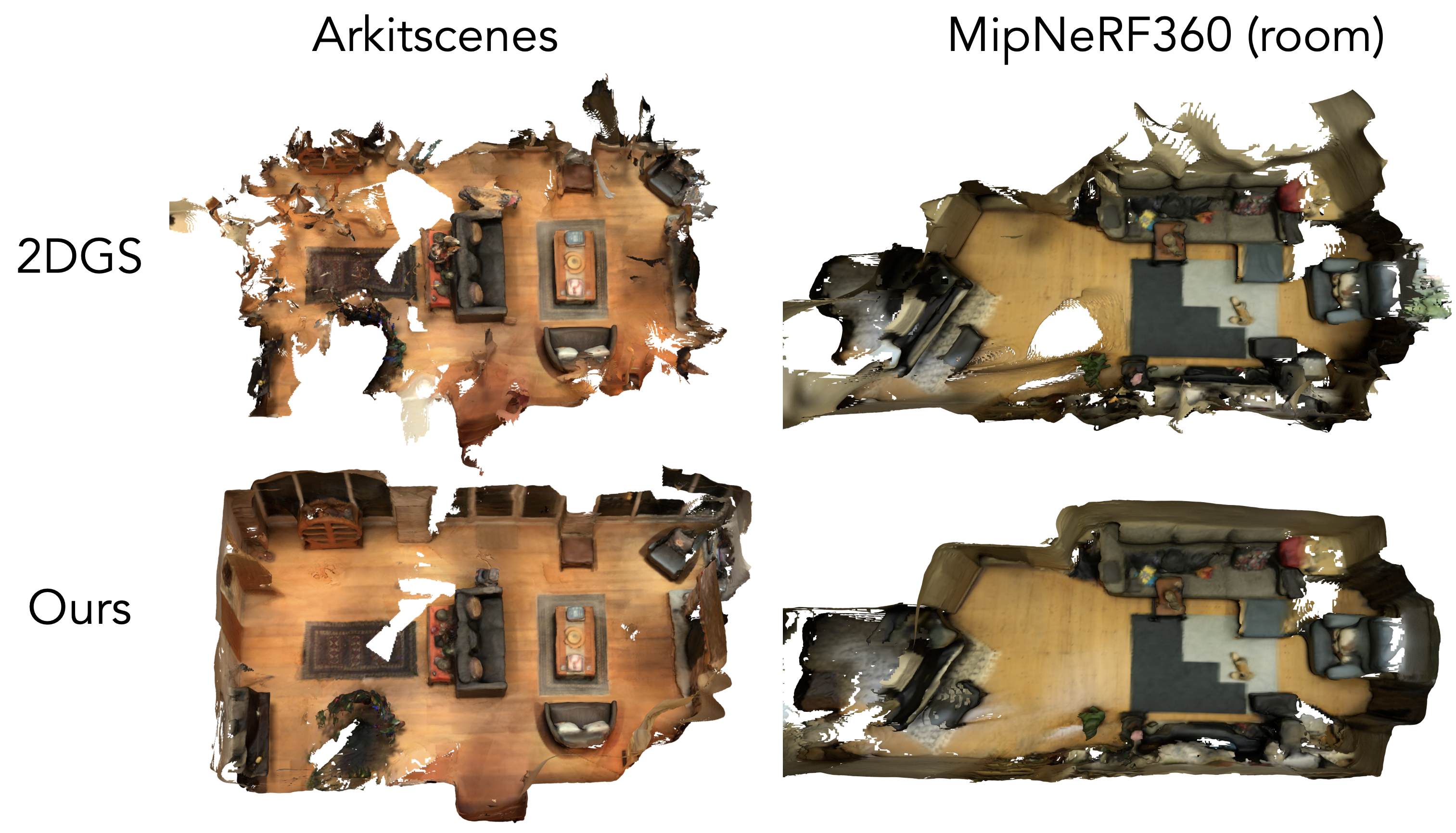}
  \vspace{-2mm}
  \caption{\textbf{Ours \vs 2DGS on ARKitScenes and MipNeRF 360.} To demonstrate the generalization ability of \OURS{}, we run our model on ARKitScenes~\cite{dehghan2021arkitscenes} and Mip-NeRF 360~\cite{barron2022mip} without fine-tuning. Compared to 2DGS, QuickSplat produces more complete geometry
  }
  \label{fig:vis-compare}
\end{figure}

\begin{figure*}[tbh]
  \centering
  \includegraphics[width=\linewidth]{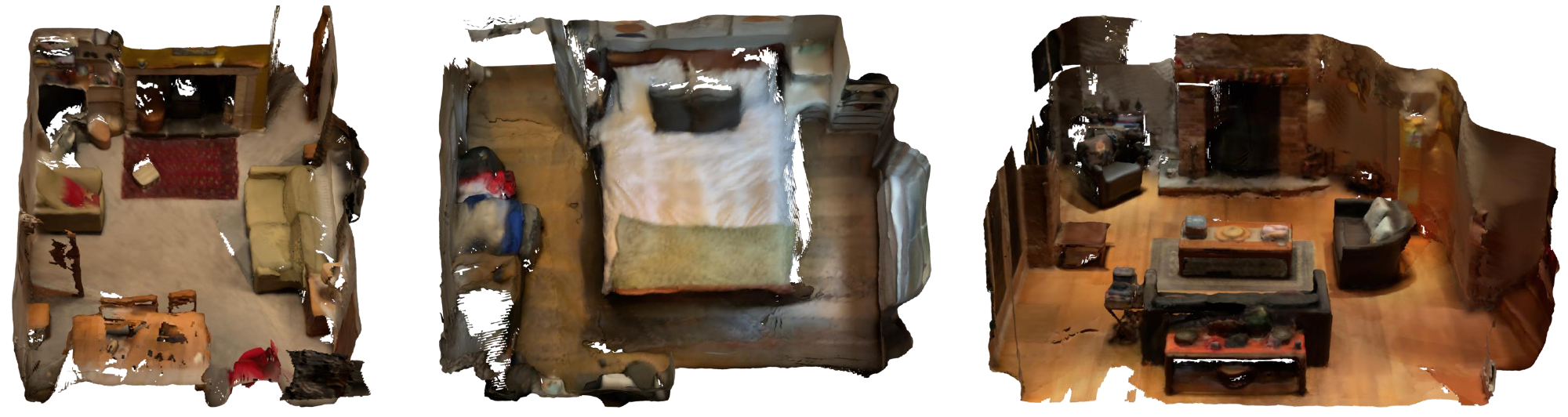}
   \caption{
    \textbf{More reconstruction result of \OURS{} on ARKitscenes dataset.} 
   }
   \label{fig:arkit-more}
\end{figure*}

This cross-dataset setting is more challenging due to the domain gap between datasets. Additionally, the RGB captures in ARKitScenes and Mip-NeRF 360 have a smaller field of view compared to ScanNet++, making reconstruction from images more difficult. We compare QuickSplat with 2DGS in \cref{tab:arkitscene} and \cref{fig:vis-compare}, which demonstrate the generalization capability of our proposed method. Additional reconstruction results are shown in \cref{fig:arkit-more}.


\begin{table}[h]
    \footnotesize
    \centering
    \begin{tabular}{@{}lcccc@{}}
    \toprule
    Method & Abs err$\downarrow$ & Acc (10cm)$\uparrow$ & Chamfer$\downarrow$ & Time \\
    \midrule
    2DGS & 0.6978 & 0.3590 & 0.6015 & 1780s\\
    Ours & \textbf{0.1775} & \textbf{0.7698} & \textbf{0.4301} & 111s \\
    \bottomrule
    \end{tabular}
    \caption{
        \textbf{Evaluation on ARKitScenes (5 scenes, no fine-tuning).}
    }
    \label{tab:arkitscene}
\end{table}

\begin{figure*}[tbh]
  \centering
  \includegraphics[width=\linewidth]{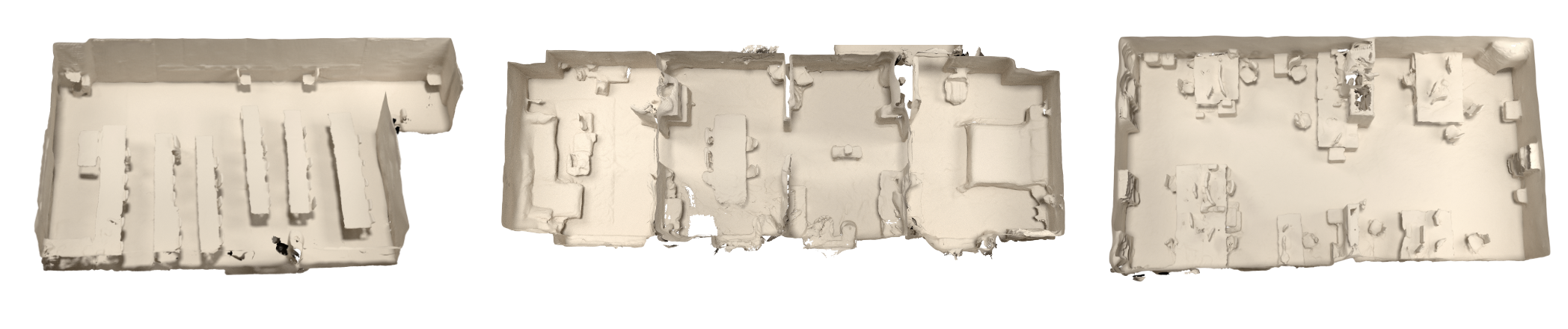}
  \vspace{-10mm}
   \caption{
    \textbf{Additional qualitative results of \OURS{} on large scenes.}
    Our method is able to reconstruct large-scale scenes, \eg, scenes containing multiple rooms, as it is not constrained by the number of the training views, and the network architecture is based on sparse convolutions.
    Even though with more training frames, \OURS{} could cost more time to optimize, it is still considerable faster than other state-of-the-arts.
   }
   \label{fig:more}
\end{figure*}

\paragraph{Large scenes.} We also demonstrate the capability to reconstruct larger scenes (\eg, indoor scenes containing multiple rooms) in \cref{fig:more}, as the method is not constrained by the number of input images. Note that the optimization times for larger scenes would increase due to the increasing number of frames during gradient accumulation. However, the overall time is still substantially faster than the existing methods.

\section{Additional Ablations}
\label{sec:more-ablations}

\paragraph{Steps} 
We ablate the number of steps for the learned optimizer and post-optimization in \cref{tab:steps}. We observe that the depth error decreases gradually over the 5 optimization steps. Additional SGD optimization steps lead to a plateau and require more time. On the other hand, the Chamfer distance changes only marginally due to the good global geometry generated by our learned initialization.

\begin{table}[h]
    \scriptsize
    \centering
    \begin{tabular}{@{}lcccccc@{}}
    \toprule
     & $T=0$ & $T=1$ & $T=2$ & $T=5$ & SGD=1k & SGD=2k \\
    \midrule
    Abs err$\downarrow$ & 0.0921 & 0.0881 & 0.0807 & 0.0732 & 0.0598 & 0.0578 \\
    Rel err$\downarrow$ & 0.0923 & 0.0792 & 0.0568 & 0.0431 & 0.0314 & 0.0292 \\
    Chamfer$\downarrow$ & 0.1478 & 0.1437 & 0.1448 & 0.1461 & 0.1361 & 0.1347 \\
    Time (s) & 0.6 & 5.7 & 11  & 26 & 77 & 124 \\
    \bottomrule
    \end{tabular}
    \vspace{-1mm}
    \caption{
        \textbf{Ablation over time steps.}
    }
    \label{tab:steps}
\end{table}

\paragraph{Optimization and densification}
We experiment with combining QuickSplat initialization with the original 2DGS optimization and densification, instead of using our optimization and densification networks, under comparable time constraints. As shown in \cref{tab:more-ablation}, the learnable optimization and densification networks achieve better reconstruction in finer details (i.e., the accuracy metrics with small thresholds). Although the original SGD optimization and densification benefit from our initialization, our full method remains more efficient.

\begin{table*}[tbh]
    \centering
    \begin{tabular}{@{}ccccccc@{}}
    \toprule
    Initializer & Optimization \& Densification & Abs err$\downarrow$ & Acc (2cm)$\uparrow$ & Acc (5cm)$\uparrow$ & Chamfer$\downarrow$ & Time \\
    \midrule
    Ours & 2DGS w/o densify & 0.0692 & 0.4650 & 0.7211 & 0.1571 & 39s \\
    Ours & 2DGS w/ densify & \textbf{0.0668} & 0.4796 & 0.7338 & 0.1486 & 39s \\
    Ours &  Ours & 0.0732 & \textbf{0.5263} & \textbf{0.7674} & \textbf{0.1461} & 26s \\
    \bottomrule
    \end{tabular}
    \vspace{-1mm}
    \caption{
        \textbf{Ablation on optimization and densification.} We compare Quicksplat's optimizer and densifier with original 2DGS optimizization (w/ and w/o densificaation) under similar time frame.
    }
    \label{tab:more-ablation}
\end{table*}

\paragraph{Extend initializer to other method}
We demonstrate that our initializer can be easily integrated into other Gaussian splatting variants, such as SAGS~\cite{ververas2024sags}. Note that we modified SAGS to use 2D Gaussians instead of 3D Gaussians as the representation for reconstructing 3D surfaces. As shown in \cref{tab:sags}, SAGS with our initialization performs significantly better than with SfM initialization. Moreover, our full method, with the learned optimization and densification, reconstructs scenes more accurately and efficiently than SAGS's original optimization and densification.

\begin{table*}[tbh]
    \centering
    \begin{tabular}{@{}ccccccc@{}}
    \toprule
    Initializer & Optimization \& Densification & Abs err$\downarrow$ & Acc (2cm)$\uparrow$ & Acc (5cm)$\uparrow$ & Chamfer$\downarrow$ & Time \\
    \midrule
    SfM & SAGS w/o densify & 0.1292 & 0.2781 & 0.5093 & 0.2879 & 429s \\
    Ours & SAGS w/o densify  & 0.0692 & 0.4724 & 0.7297 & 0.1633 & 253s \\
    Ours & SAGS w/ densify  & \textbf{0.0669} & 0.4825 & 0.7381 & 0.1625 & 276s \\
    Ours & Ours & 0.0732 & \textbf{0.5263} & \textbf{0.7674} & \textbf{0.1461} & 26s \\
    \bottomrule
    \end{tabular}
    \vspace{-1mm}
    \caption{
        \textbf{Combined with SAGS \cite{ververas2024sags}.} 
        We show that our initializer can be easily integrated into other methods, resulting in improved performance. In addition, our learned densification and optimization are faster and more accurate than SAGS under the same initialization. (Note that we modified SAGS to output 2D Gaussian splats for surface reconstruction.)
    }
    \label{tab:sags}
\end{table*}

\end{appendices}
\end{document}